\pdfoutput=1

\documentclass[11pt]{article}

\usepackage[]{acl}

\usepackage{times}
\usepackage{latexsym}

\usepackage[T1]{fontenc}

\usepackage[utf8]{inputenc}

\usepackage{microtype}

\usepackage{graphicx}
\usepackage{eurosym}
\usepackage{tablefootnote}
\usepackage{kantlipsum}
\usepackage{tabularx}
\graphicspath{ {images/} }
\usepackage{amsmath}
\usepackage{hhline}
\usepackage[utf8]{inputenc}
\usepackage{multirow}

\usepackage[T1]{fontenc}
\usepackage{kantlipsum}
\usepackage{listings}
\usepackage{stfloats}
\usepackage{csquotes}
\usepackage{array}
\usepackage{booktabs}
\usepackage[super]{nth}
\usepackage[inline,shortlabels]{enumitem}
\usepackage{enumitem}
\usepackage{makecell}
\usepackage{siunitx}
\usepackage{xcolor, colortbl}
\usepackage{bm}
\usepackage{url}
\usepackage{wrapfig}
\usepackage{amsmath}
\usepackage{amssymb}
\usepackage{mathtools}
\usepackage{amsfonts}
\usepackage{enumitem}
\usepackage[ruled,linesnumbered]{algorithm2e}
\usepackage{graphicx}
\usepackage{subcaption}
\usepackage{nicefrac}       
\usepackage{tablefootnote}
\usepackage{bbm}
\usepackage{microtype} 
\usepackage{dsfont}
\usepackage[ruled,linesnumbered]{algorithm2e}

\newcommand{\bb}{\textsc{BatchBALD}}
\newcommand{\bald}{\textsc{BALD}}
\newcommand{\entropy}{\textsc{Entropy}}
\newcommand{\lc}{\textsc{Least Confidence}}
\newcommand{\rand}{\textsc{Random}}
\newcommand{\alps}{\textsc{Alps}}
\newcommand{\badge}{\textsc{Badge}}
\newcommand{\bkm}{\textsc{BertKM}}

\newcommand{\bert}{\textsc{Bert}}
\newcommand{\tapt}{\textsc{tapt}}

\newcommand{\Dpool}{\mathcal{D}_{\textbf{pool}}}
\newcommand{\Dlab}{\mathcal{D}_{\textbf{lab}}}
\newcommand{\Dval}{\mathcal{D}_{\textbf{val}}}
\newcommand{\Dtest}{\mathcal{D}_{\textbf{test}}}

\newcommand{\trec}{\textsc{trec-6}}
\newcommand{\imdb}{\textsc{imdb}}
\newcommand{\sst}{\textsc{sst-2}}

\newcommand{\dbpedia}{\textsc{dbpedia}}
\newcommand{\ag}{\textsc{agnews}}

\newcommand{\sft}{\textsc{sft}}
\newcommand{\ft}{\textsc{ft+}}

\title{On the Importance of Effectively Adapting Pretrained Language Models for Active Learning}

\author{Katerina Margatina$^\spadesuit$ \enspace
Lo\"{i}c Barrault$^\clubsuit$ \enspace Nikolaos Aletras$^\spadesuit$ \\
    $^\spadesuit$University of Sheffield, $^\clubsuit$University of Le Mans \\
  \texttt{\{k.margatina,n.aletras\}@sheffield.ac.uk}\\ \texttt{loic.barrault@univ-lemans.fr}
  }

\begin{document}
\maketitle
\begin{abstract}
Recent Active Learning (AL) approaches in Natural Language Processing (NLP) proposed using off-the-shelf pretrained language models (LMs). In this paper, we argue that these LMs are not adapted effectively to the downstream task during AL and we explore ways to address this issue. We suggest to first adapt the pretrained LM to the target task by continuing training with all the available \textit{unlabeled} data and then use it for AL. We also propose a simple yet effective fine-tuning method to ensure that the adapted LM is properly trained in both low and high resource scenarios during AL. Our experiments demonstrate that our approach provides substantial data efficiency improvements compared to the standard fine-tuning approach, suggesting that a poor training strategy can be catastrophic for AL.\footnote{For all experiments in this paper, we have used the code provided by \citet{margatina-etal-2021-active}: \url{https://github.com/mourga/contrastive-active-learning}}
\end{abstract}

\section{Introduction}\label{sec:intro}
Active Learning (AL) is a method for training supervised models in a data-efficient way~\cite{Cohn:1996:ALS:1622737.1622744, settles2009active}.  It is especially useful in scenarios where a large pool of unlabeled data is available but only a limited annotation budget can be afforded; or where expert annotation is prohibitively expensive and time consuming. AL methods iteratively alternate between (i) model training with the labeled data available; and (ii) data selection for annotation using a stopping criterion, e.g. until exhausting a fixed annotation budget or reaching a pre-defined performance on a held-out dataset. 

Data selection is performed by an acquisition function that ranks unlabeled data points by some \textit{informativeness} metric aiming to improve over random selection, using either uncertainty~\cite{Lewis:1994:SAT:188490.188495,Cohn:1996:ALS:1622737.1622744,Gal2017-gh, Kirsch2019-lk, zhang-plank-2021-cartography-active}, diversity~\cite{Brinker03incorporatingdiversity, pmlr-v16-bodo11a,conf/iclr/SenerS18}, or both~\cite{Ducoffe2018-sq, Ash2020Deep,yuan2020coldstart, margatina-etal-2021-active}.

Previous AL approaches in NLP use task-specific neural models that are trained from scratch at each iteration~\cite{Shen2017-km, Siddhant2018-lg, Prabhu2019-hm, ikhwantri-etal-2018-multi, Kasai2019-ai}. However, these models are usually outperformed by pretrained language models (LMs) adapted to end-tasks~\cite{Howard2018-vf}, making them suboptimal for AL. Only recently, pretrained LMs such as \bert{}~\cite{Devlin2019-ou} have been introduced in AL settings~\cite{yuan2020coldstart, Ein-Dor2020-mm,shelmanov-etal-2021-active, karamcheti-etal-2021-mind, margatina-etal-2021-active}. Still, they are trained at each AL iteration with a standard fine-tuning approach that mainly includes a pre-defined number of training epochs, which has been demonstrated to be unstable, especially in small datasets~\cite{Zhang2020RevisitingFB, Dodge2020FineTuningPL,mosbach2021on}. Since AL includes both low and high data resource settings, the AL model training scheme should be robust in both scenarios.\footnote{During the first few AL iterations the available labeled data is limited (\textit{low-resource}), while it could become very large towards the last iterations (\textit{high-resource}).} 

To address these limitations, we introduce a suite of effective training strategies for AL (\S\ref{sec:balm}). Contrary to previous work~\cite{yuan2020coldstart, Ein-Dor2020-mm,margatina-etal-2021-active} that also use \bert{}~\cite{Devlin2019-ou}, our proposed method accounts for various data availability settings and the instability of fine-tuning. First, we continue \textit{pretraining} the LM with the available \textit{unlabeled} data to adapt it to the task-specific domain. This way, we leverage not only the available labeled data at each AL iteration, but the entire unlabeled pool. Second, we further propose a simple yet effective fine-tuning method that is robust in both low and high resource data settings for AL.

We explore the effectiveness of our approach on five standard natural language understandings tasks with various acquisition functions, showing that it outperforms all baselines (\S\ref{sec:results}). We also conduct an analysis to demonstrate the importance of effective adaptation of pretrained models for AL (\S\ref{sec:analysis}). Our findings highlight that the LM adaptation strategy can be more critical than the actual data acquisition strategy. 

\section{Adapting \& Fine-tuning Pretrained Models for Active Learning}\label{sec:balm}
Given a downstream classification task with $C$ classes, a typical AL setup consists of a pool of unlabeled data $\Dpool$, a model $\mathcal{M}$, an annotation budget $b$ of data points and an acquisition function $a(.)$ for selecting $k$ unlabeled data points for annotation (i.e. acquisition size) until $b$ runs out. The AL performance is assessed by training a model on the actively acquired dataset and evaluating on a held-out test set $\Dtest$.

\paragraph{Adaptation (\tapt{})}\label{sec:balm_tapt}
Inspired by recent work on transfer learning that shows improvements in downstream classification performance by continuing the pretraining of the LM with the task data~\cite{Howard2018-vf} we add an extra step to the AL process by continuing pretraining the LM (i.e. Task-Adaptive Pretraining \tapt), as in \citet{gururangan-etal-2020-dont}. Formally, we use an LM, such as \bert~\cite{Devlin2019-ou}, $\mathcal{P}(x;W_0)$  with weights $W_0$, that has been already pretrained on a large corpus. We fine-tune $\mathcal{P}(x;W_0)$ with the available unlabeled data of the downstream task $\Dpool$, resulting in the task-adapted LM $\mathcal{P}_{\text{TAPT}}(x;W_0')$ with new weights $W_0'$ (cf. line 2 of \autoref{algo:balm}).

\paragraph{Fine-tuning (\ft{})}\label{sec:balm_ft}
We now use the adapted LM $\mathcal{P}_{\text{TAPT}}(x;W_0')$ for AL.
At each iteration $i$, we initialize our model $\mathcal{M}_i$ with the pretrained weights $W_0'$ and we add a task-specific feedforward layer for classification with weights $W_c$ on top of the \texttt{[CLS]} token representation of \bert{}-based $\mathcal{P}_{\text{TAPT}}$.
We fine-tune the classification model $\mathcal{M}_i(x;[W_0',W_c])$ with all $x \in \Dlab$. (cf. line 6 to 8 of \autoref{algo:balm}).
 
Recent work in AL~\cite{Ein-Dor2020-mm, yuan2020coldstart} uses the standard fine-tuning method proposed in \citet{Devlin2019-ou} which includes a fixed number of $3$ training epochs, learning rate warmup over the first $10\%$ of the steps and AdamW optimizer~\cite{loshchilov2018decoupled} without bias correction, among other hyperparameters.

We follow a different approach by taking into account insights from few-shot fine-tuning literature~\cite{mosbach2021on, Zhang2020RevisitingFB,Dodge2020FineTuningPL} that proposes longer fine-tuning and more evaluation steps during training.~\footnote{In this paper we use \textit{few-shot} to describe the setting where there are \textit{few} labeled data available and therefore \textit{few-shot fine-tuning} corresponds to fine-tuning a model on limited labeled training data. This is different than the few-shot setting presented in recent literature \cite{NEURIPS2020_1457c0d6}, where no model weights are updated. }
We combine these guidelines to our fine-tuning approach by using early stopping with $20$ epochs based on the validation loss, learning rate $2e-5$, bias correction and $5$ evaluation steps per epoch. 
However, increasing the number of epochs from $3$ to $20$, also increases the warmup steps ($10\%$ of total steps\footnote{Some guidelines propose an even smaller number of warmup steps, such as $6\%$ in RoBERTa~\cite{liu2020roberta}.}) almost $7$ times. This may be problematic in scenarios where the dataset is large but the optimal number of epochs may be small (e.g. $2$ or $3$). To account for this limitation in our AL setting where the size of training set changes at each iteration, we propose
to select the warmup steps as $min(10\% \text{ of total steps}, 100)$. We denote standard fine-tuning as \sft{} and our approach as \ft{}.

\begin{algorithm}[!t]

	\caption{AL with Pretrained LMs \label{algo:balm}}

    \DontPrintSemicolon
    \SetAlgoLined

	\KwIn{unlabeled data $\Dpool$, pretrained LM $\mathcal{P}(x;W_0)$, acquisition size $k$, AL iterations $T$, acquisition function $a$}
    
    $\Dlab \leftarrow \emptyset$ \;
    
    $\mathcal{P}_{\text{TAPT}}(x;W_0') \leftarrow$ Train $\mathcal{P}(x;W_0)$ on $\Dpool$ \;
    
    $\mathcal{Q}_0 \leftarrow $\ RANDOM$(.),  |\mathcal{Q}_0|=k$ \;
	
	$\Dlab = \Dlab \cup \mathcal{Q}_0$ \;
	$\Dpool = \Dpool \setminus \mathcal{Q}_0$ \;
	\For { $i \leftarrow 1$ \KwTo $T$}{
	$\mathcal{M}_i(x;[W_0',W_c]) \leftarrow$ Initialize from $\mathcal{P}_{\text{TAPT}}(x;W_0')$ \;
	$\mathcal{M}_i(x;W_i) \leftarrow$ Train model on $\Dlab$ \;
	$\mathcal{Q}_i \leftarrow a(\mathcal{M}_i, \Dpool, k)$ \;
	$\Dlab = \Dlab \cup \mathcal{Q}_i$ \;
	$\Dpool = \Dpool \setminus \mathcal{Q}_i$ \;
	}
	\KwOut{ $\Dlab$ }
\end{algorithm}


\setlength{\tabcolsep}{6pt} 
\renewcommand{\arraystretch}{1.1} 

\begin{table}[!t]
\centering
\resizebox{0.95\columnwidth}{!}{%
\centering
\begin{tabular}{lccccc}
\Xhline{2\arrayrulewidth}

\textsc{datasets} & \textsc{train} & \textsc{val}  & \textsc{test} &  $k$ & $C$\\\hline 
\textsc{trec-6}  & $4.9$K & $546$ & $500$ &  $1\%$ & $6$\\ 
\textsc{dbpedia} & $20$K  & $2$K & $70$K &  $1$\% & $14$\\
\textsc{imdb}  & $22.5$K & $2.5$K & $25$K&  $1$\% & $2$\\
\textsc{sst-2} & $60.6$K  & $6.7$K & $871$&  $1$\% & $2$\\
\textsc{agnews} & $114$K  & $6$K & $7.6$K & $0.5\%$ & $4$\\

\Xhline{2\arrayrulewidth}
\end{tabular}
}

\caption{Datasets statistics for
$\Dpool$, $\Dval$ and $\Dtest$ respectively. $k$ stands for the acquisition size (\% of $\Dpool$) and $C$ the number of classes.}

\label{table:datasets}

\end{table}

\begin{figure*}[!t]
\centering
    \begin{subfigure}{0.32\textwidth}
        \centering
        \includegraphics[width=\textwidth]{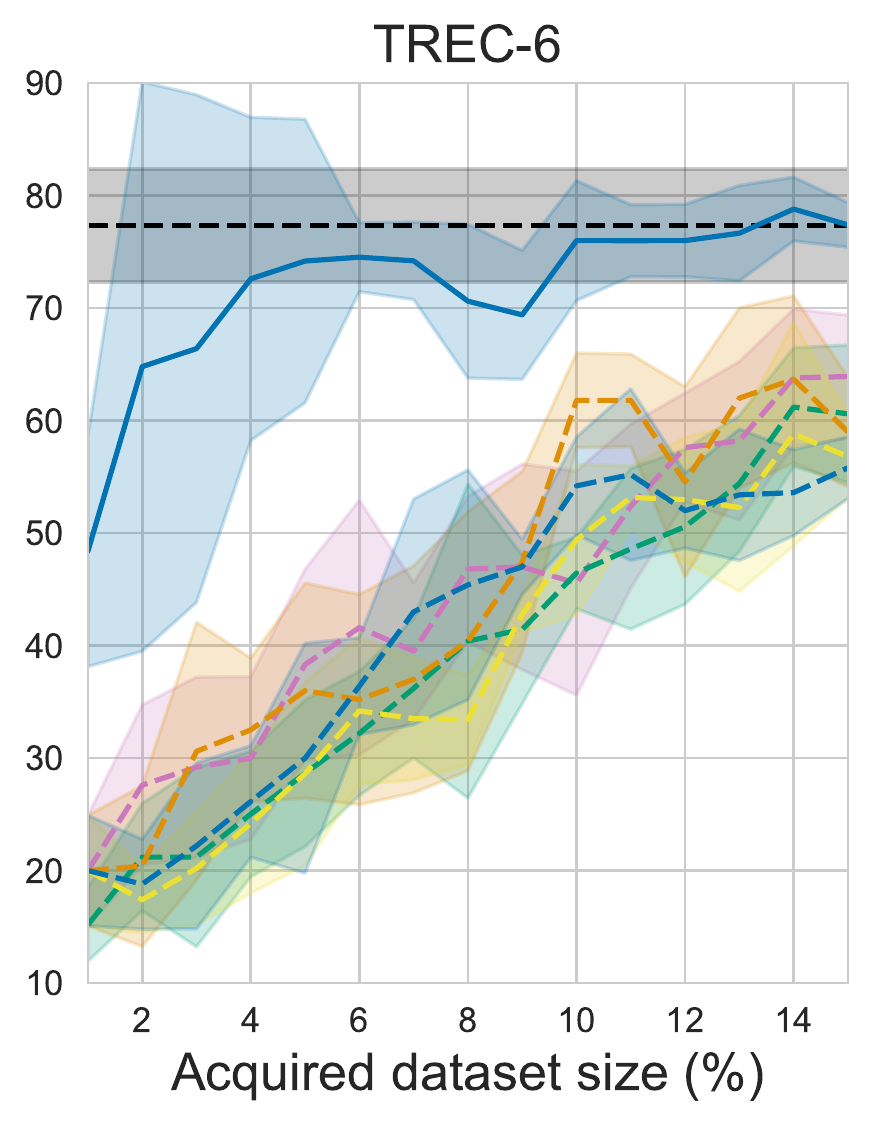}
    \end{subfigure}%
    \begin{subfigure}{0.32\textwidth}
        \centering
        \includegraphics[width=\textwidth]{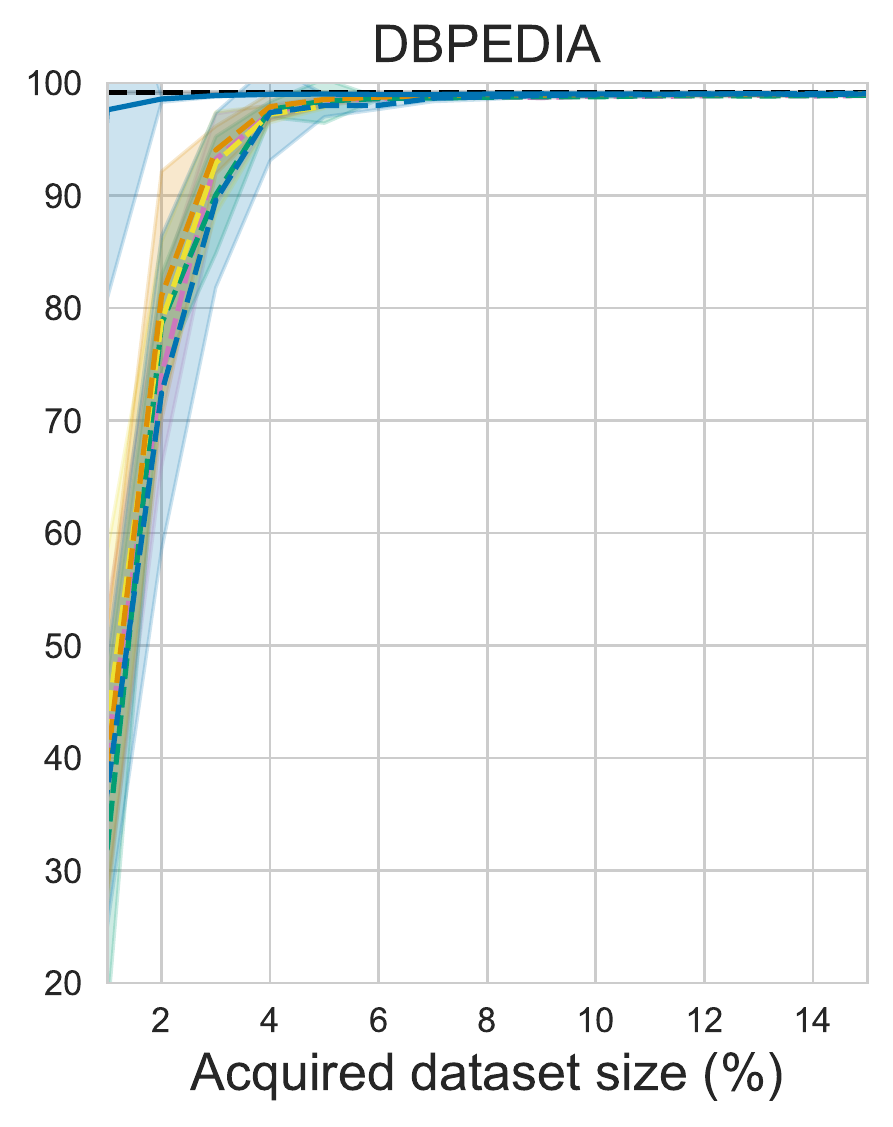}
    \end{subfigure}%
    \begin{subfigure}{0.32\textwidth}
        \centering
        \includegraphics[width=\textwidth]{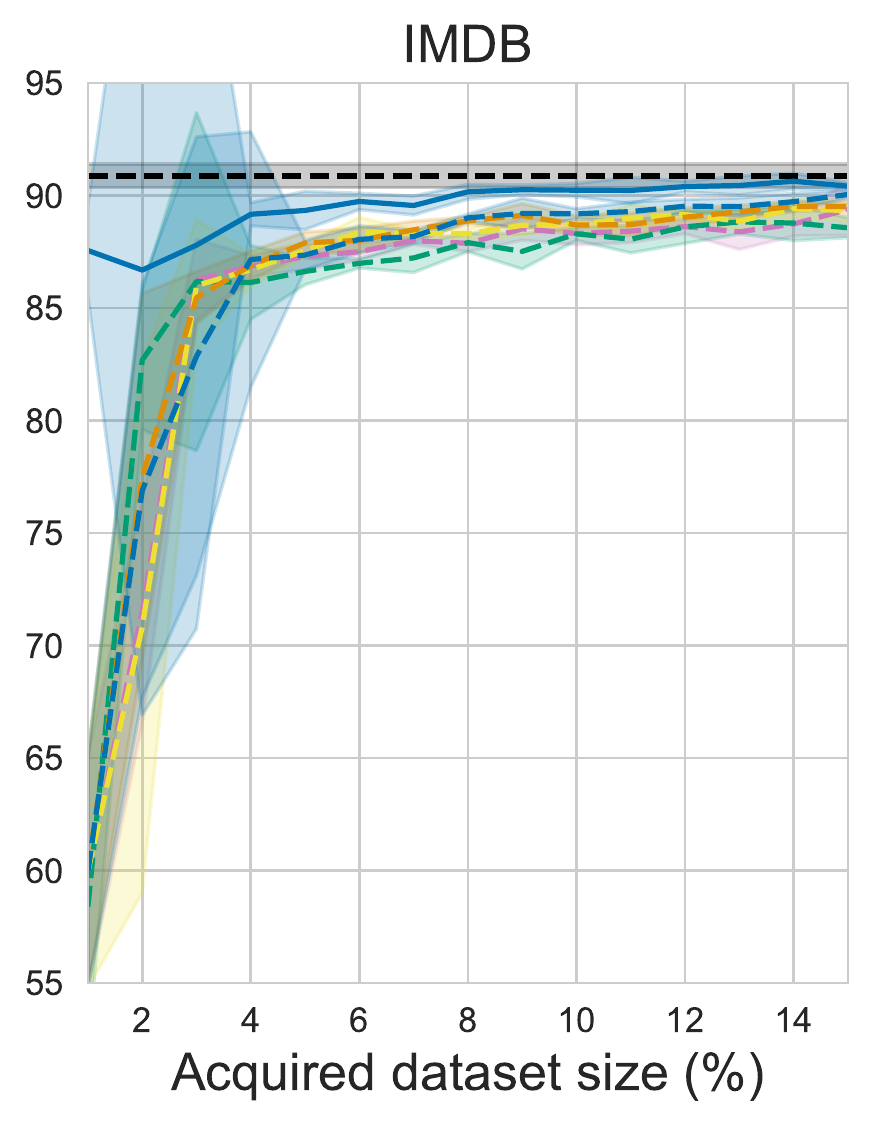}
    \end{subfigure}%
    \\
    \begin{subfigure}{0.32\textwidth}
        \centering
        \includegraphics[width=\textwidth]{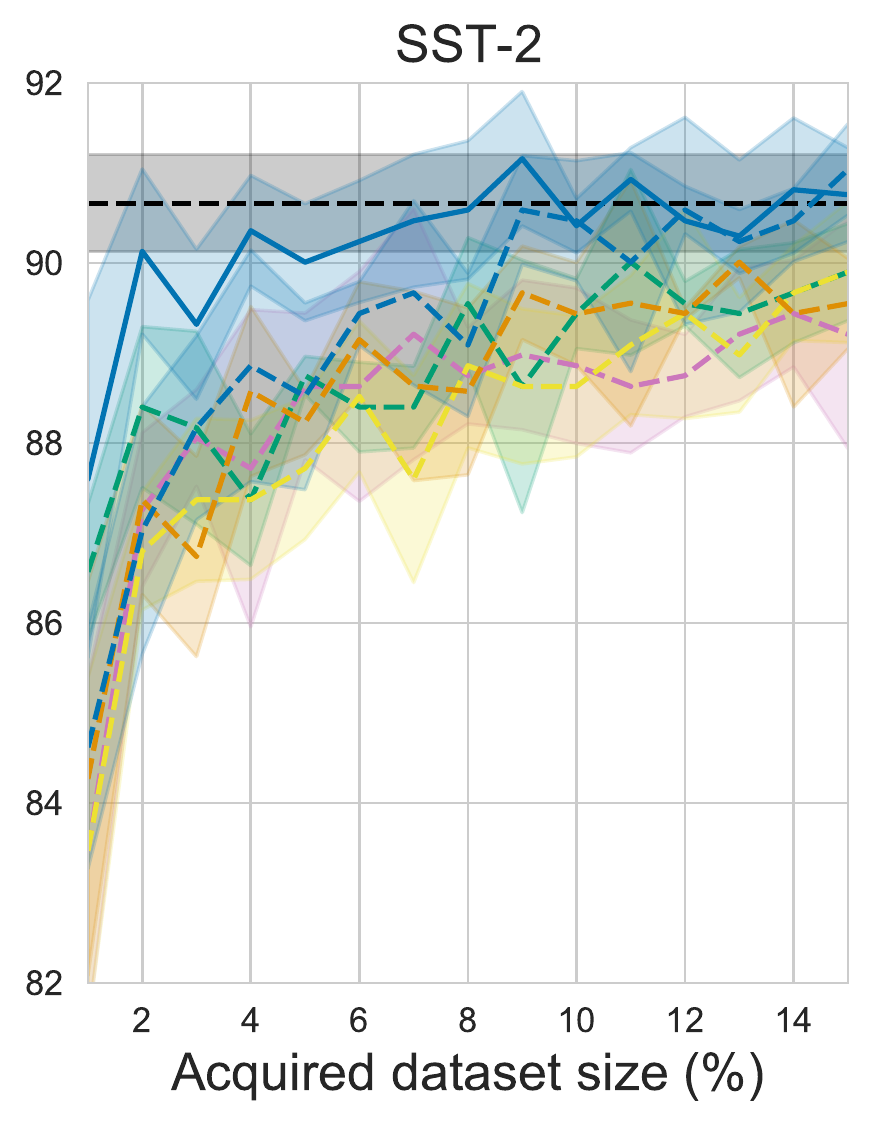}
        \end{subfigure}%
    \begin{subfigure}{0.32\textwidth}
        \includegraphics[width=\textwidth]{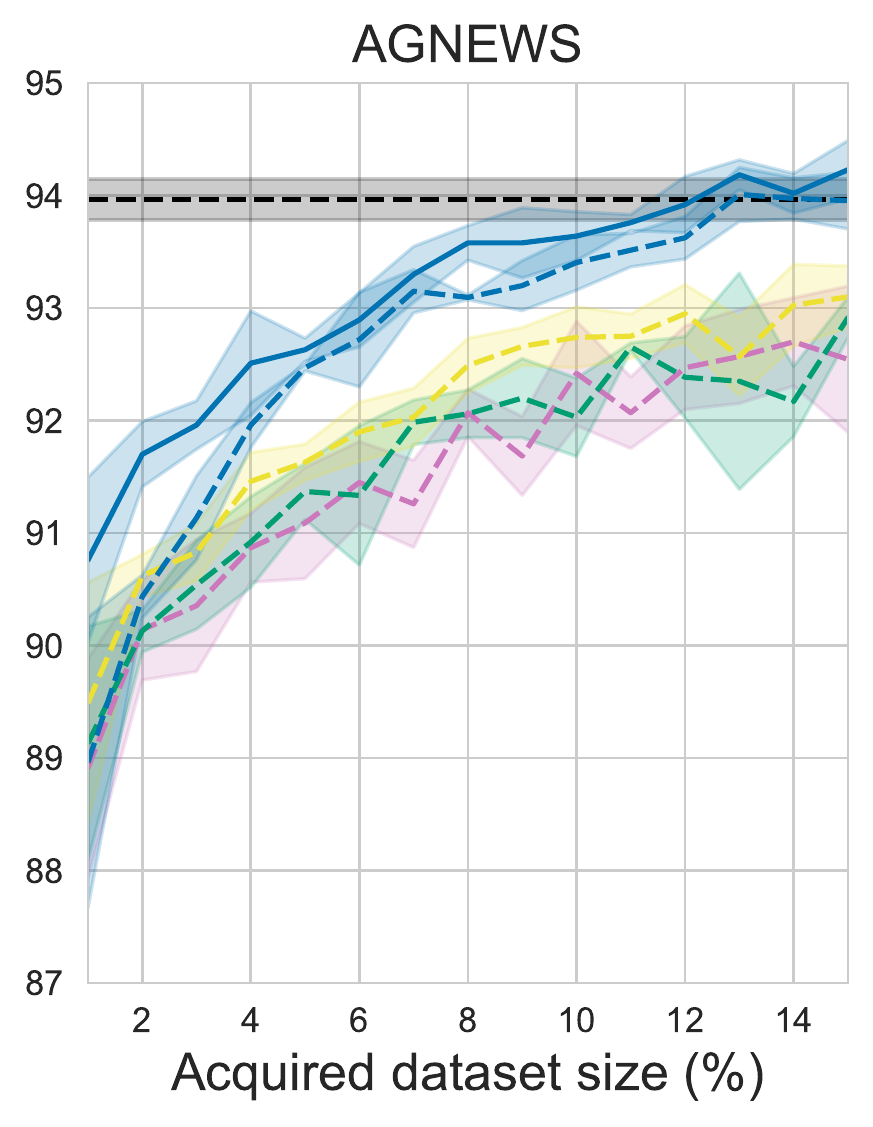}
            \end{subfigure}%
        \begin{subfigure}{0.32\textwidth}
        \centering
        \includegraphics[width=0.7\textwidth]{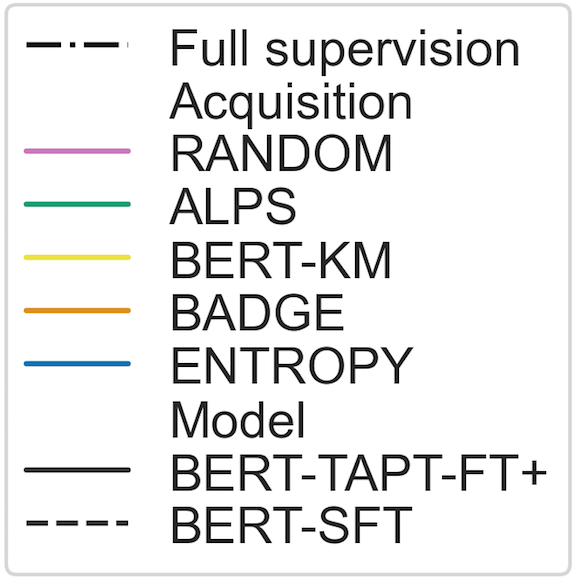}
    \end{subfigure}%
    \caption{Test accuracy during AL iterations.
    We plot the median and standard deviation across five runs. 
    }
    \label{fig:al}
\end{figure*}
\section{Experiments \& Results}\label{sec:results}

\paragraph{Data}
We experiment with five 
diverse 
natural language understanding tasks: question classification (\trec{};~\citet{article-trec}), sentiment analysis (\imdb;~\citet{maas-etal-2011-learning}, \sst~\citet{socher-etal-2013-recursive}) and topic classification (\dbpedia, \ag{};~\citet{NIPS2015_250cf8b5}),
including binary and multi-class labels and varying dataset sizes (Table~\ref{table:datasets}). More details can be found in Appendix~\ref{sec:datasets}.

\paragraph{Experimental Setup}
We perform all AL experiments using \texttt{BERT-base}~\cite{Devlin2019-ou} and \entropy{}, \bkm{}, \alps{}~\cite{yuan2020coldstart}, \badge{}~\cite{Ash2020Deep} and \rand{} (baseline) as the acquisition functions. We pair our proposed training approach \tapt{}-\ft{} with \entropy{} acquisition. We refer the reader to Appendix~\ref{sec:appendix_exp_setup} for an extended description of our experimental setup, including the datasets used (\S\ref{sec:datasets}), the training and AL details (\S\ref{sec:training_details}), the model hyperparameters (\S\ref{sec:hyperparameters}) and the baselines (\S\ref{sec:baselines}).

\paragraph{Results} Figure~\ref{fig:al} shows the test accuracy during AL iterations. We first observe that our proposed approach (\tapt{}-\ft{}) 
achieves large data efficiency reaching the full-dataset performance within the $15\%$ budget for all datasets, in contrast to the standard AL approach (\bert-\sft{}).
The effectiveness of our approach is mostly notable in the smaller datasets. In \trec{}, it achieves the goal accuracy with almost $10\%$ annotated data, while in \dbpedia{} only in the first iteration with $2\%$ of the data. After the first AL iteration in \imdb{}, \tapt{}-\ft{}, it achieves only  $2.5$ points of accuracy lower than the performance when using $100\%$ of the data.
In the larger \sst{} and \ag{} datasets, it is closer to the baselines but still outperforms them, achieving the full-dataset performance with $8\%$ and $12\%$ of the data respectively.
We also observe that in all 
five 
datasets, the addition of our proposed pretraining step (\tapt{}) and fine-tuning technique (\ft{}) leads to large performance gains, 
especially in the first AL iterations. This is particularly evident in \trec{}, \dbpedia{} and \imdb{} datasets, where after the \textit{first} AL iteration (i.e. equivalent to $2\%$ of training data)
\tapt{}+\ft{} with \entropy{} is $45$, $30$ and $12$ points in accuracy higher than the \entropy{} baseline with \bert{} and \sft{}. 

\paragraph{Training vs. Acquisition Strategy} We finally observe that the performance curves of the various acquisition functions considered (i.e. dotted lines) are generally close to each other, suggesting that the choice of the acquisition strategy may not affect substantially the AL performance in certain cases. In other words, we conclude that \textit{the training strategy can be more important than the acquisition strategy}. We find that uncertainty sampling with \entropy{} is generally the best performing acquisition function, followed by \badge{}.\footnote{We provide results with additional acquisition functions in the Appendix~\ref{sec:perf_afs} and~\ref{sec:appendix_acquisition}.} Still, finding a universally well-performing acquisition function, independent of the training strategy, is an open research question. 

\section{Analysis \& Discussion}\label{sec:analysis}
\subsection{Task-Adaptive Pretraining}\label{sec:analysis_tapt}
\begin{figure}[!t]
    \resizebox{\columnwidth}{!}{%
        \centering
        \includegraphics[width=\textwidth]{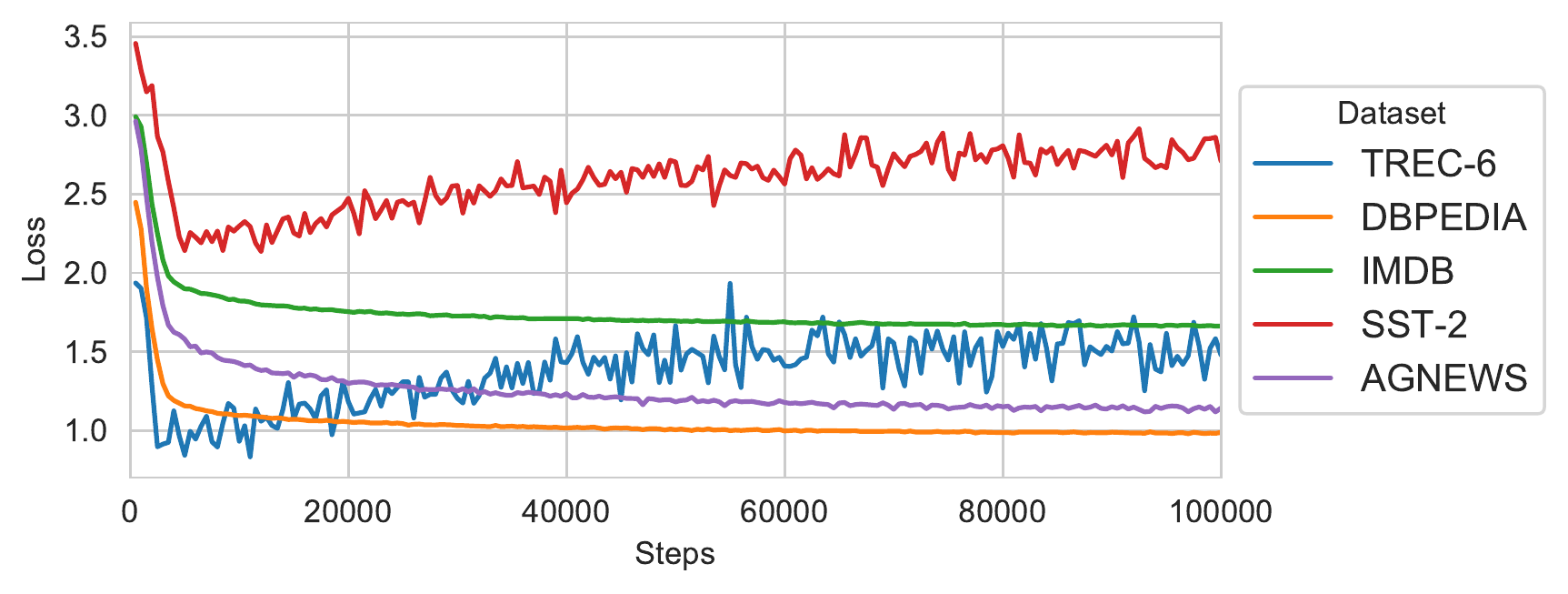}
        }
    \caption{Validation MLM loss during \tapt{}.}
    \label{fig:tapt_loss}
\end{figure}

We first present details of 
our implementation of 
\tapt{} (\S\ref{sec:balm_tapt}) and reflect on its effectiveness in the AL pipeline. Following \citet{gururangan-etal-2020-dont}, we continue pretraining \bert{}
for the MLM task using all the unlabeled data $\Dpool$ 
for all datasets separately.
We plot the learning curves of \bert-\tapt{} for all datasets in Figure~\ref{fig:tapt_loss}. 
We first observe that the masked LM 
loss is steadily decreasing for \dbpedia{}, \imdb{} and \ag{} across optimization steps, which correlates with the high early AL performance gains of \tapt{} in these datasets (Fig.~\ref{fig:al}).
We also observe that the LM overfits in \trec{} and \sst{} datasets. We attribute this to the very small training dataset of \trec{} and the informal textual style of \sst{}. 
Despite the fact that 
the 
\sst{} 
dataset 
includes approximately $67$K of training data, the sentences are very short (i.e. average length of $9.4$ words per sentence). We hypothesize the LM overfits because of the lack of long and more diverse sentences.
We provide more details on \tapt{} at the Appendix~\ref{sec:appendix_tapt}.

\begin{figure}[t!]
    \resizebox{\columnwidth}{!}{%
    \begin{subfigure}{0.25\textwidth}
        \centering
        \includegraphics[width=\textwidth]{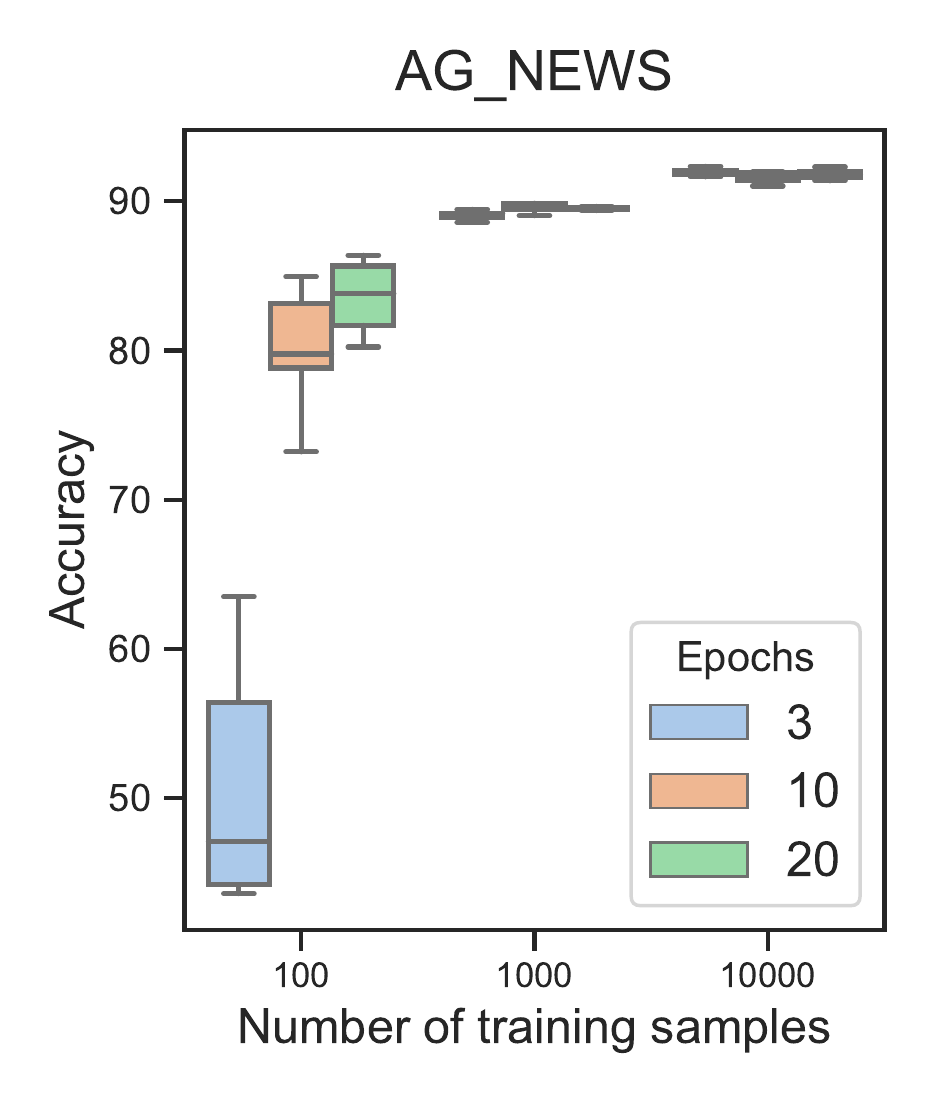}
    \end{subfigure}%
    \begin{subfigure}{0.25\textwidth}
        \centering
        \includegraphics[width=\textwidth]{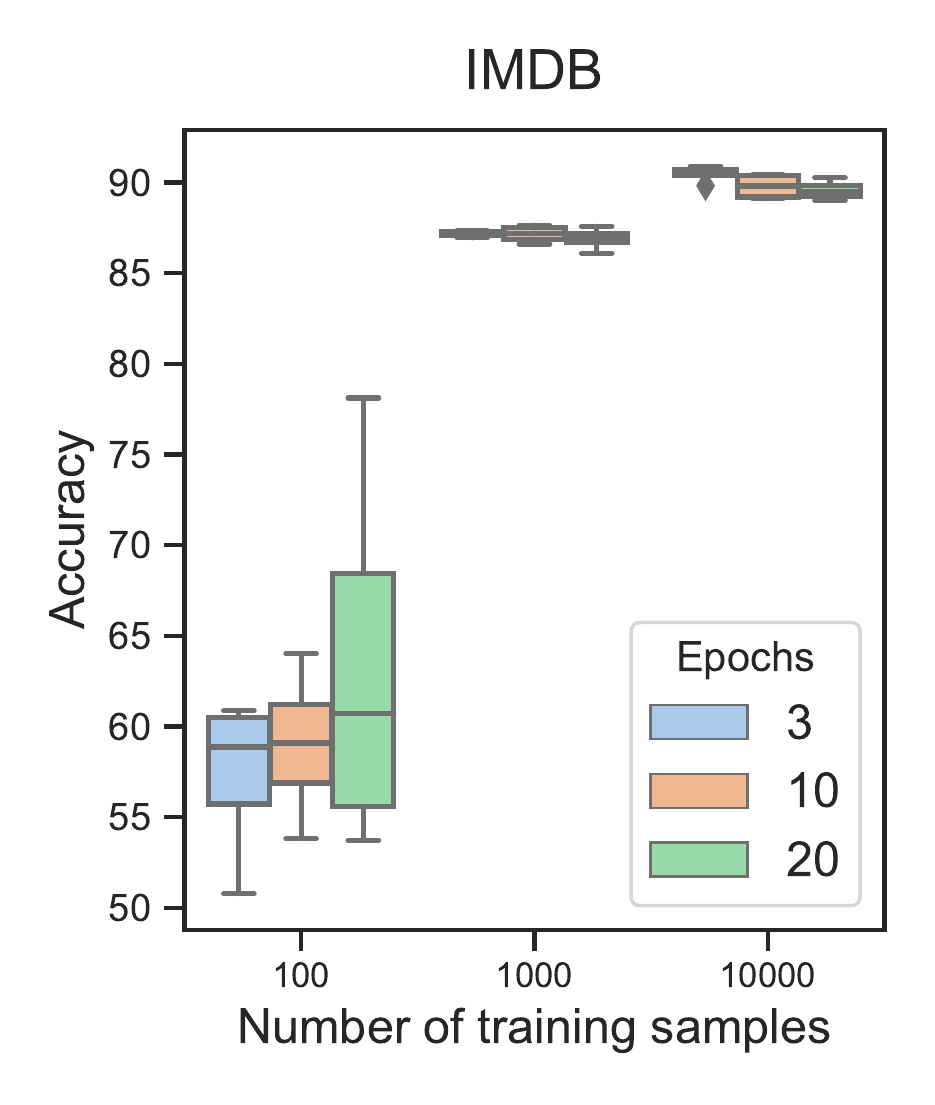}
    \end{subfigure}%
    }
    \caption{Few-shot standard \bert{} fine-tuning.}
    \label{fig:few_shot}
\end{figure}
\subsection{Few-shot Fine-tuning}\label{sec:analysis_ft}
In this set of experiments, we aim to highlight that it is crucial to consider the few-shot learning problem in the early AL stages,
which is often neglected in literature. This is more important when using pretrained LMs, since they are overparameterized models that require adapting their training scheme in low data settings to ensure robustness.

To illustrate the potential ineffectiveness of standard fine-tuning (\sft{}), we randomly undersample the \ag{} and \imdb{} datasets
to form low, medium and high resource data settings 
(i.e. $100$, $1,000$ and $10,000$ training samples),
and train \bert{} for a fixed number of $3$, $10$, and $20$ epochs. 
We repeat this process with $10$ different random seeds to account for stochasticity in sampling and we plot the test accuracy in Figure~\ref{fig:few_shot}.
Figure~\ref{fig:few_shot} shows that \sft{} is suboptimal for low data settings
(e.g. $100$ samples), 
indicating that more optimization steps 
(i.e. epochs) 
are needed for the model to adapt to the few training samples~\cite{Zhang2020RevisitingFB,mosbach2021on}. As the training samples increase
(e.g. $1,000$),
fewer epochs are often better. It is thus evident that there is not a
clearly 
optimal way to choose a predefined number of epochs to train the model given the number of training examples. This motivates the need to find a fine-tuning policy for AL that effectively adapts to the data resource setting of each iteration
(independently of the number of training examples or dataset), 
which is mainly tackled by our proposed fine-tuning approach \ft{} (\S\ref{sec:balm_ft}).

\subsection{Ablation Study}\label{sec:ablation}
We finally conduct an ablation study to evaluate the contribution of our two proposed steps to the AL pipeline; the pretraining step (\tapt{}) and 
fine-tuning method (\ft{}). We show that the addition of both methods provides large gains compared to standard fine-tuning (\sft{}) 
in terms of accuracy, data efficiency and uncertainty calibration.
We compare \bert{} with \sft{}, \bert{} with \ft{} and \bert-\tapt{} with \ft{}. Along with test accuracy, we also evaluate each model using uncertainty estimation metrics~\cite{NEURIPS2019_8558cb40}: Brier score, negative log likelihood (NLL), expected calibration error (ECE) and entropy.
A well-calibrated model should have high accuracy and low uncertainty. 

\begin{figure}[!t]
\centering
    \begin{subfigure}{0.48\columnwidth}
        \centering
        \includegraphics[width=\textwidth]{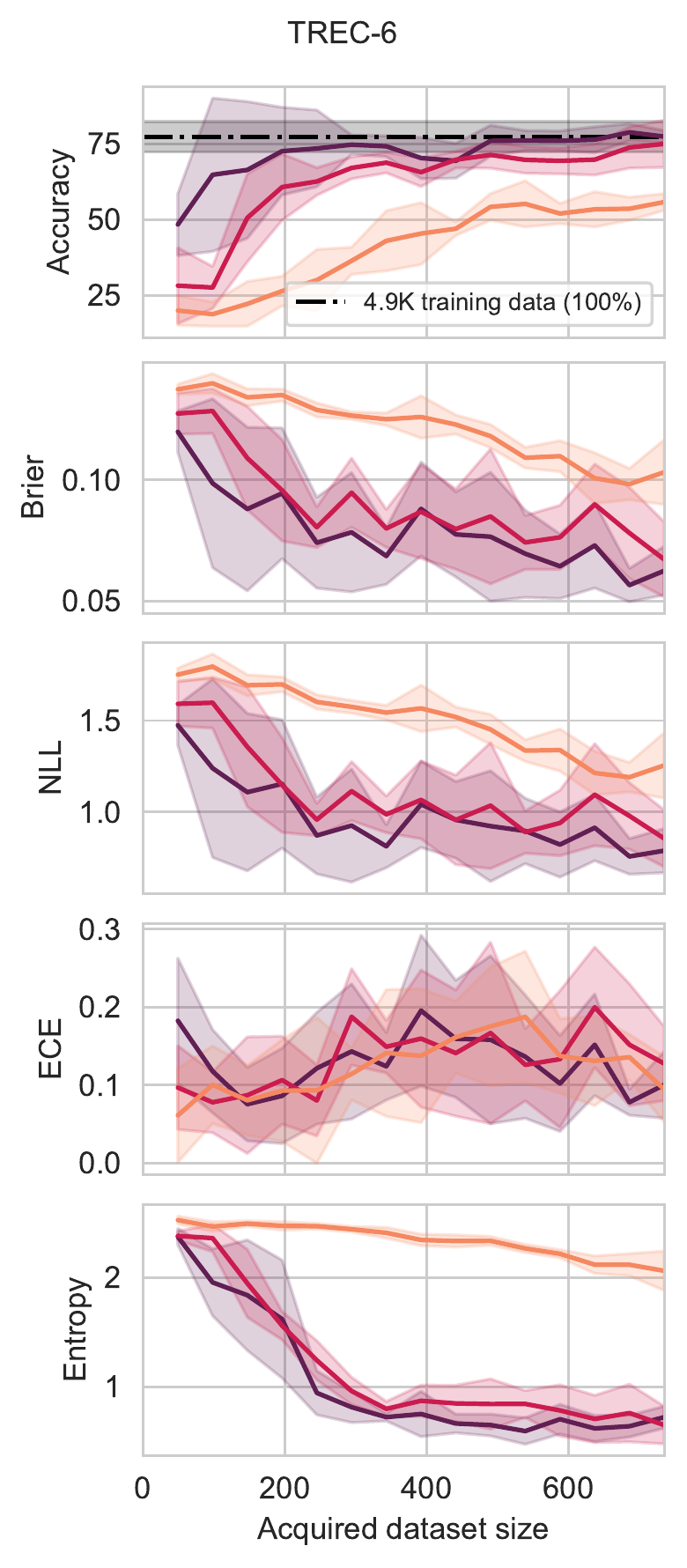}
    \end{subfigure}%
    \hfill
    \begin{subfigure}{0.48\columnwidth}
        \centering
        \includegraphics[width=\textwidth]{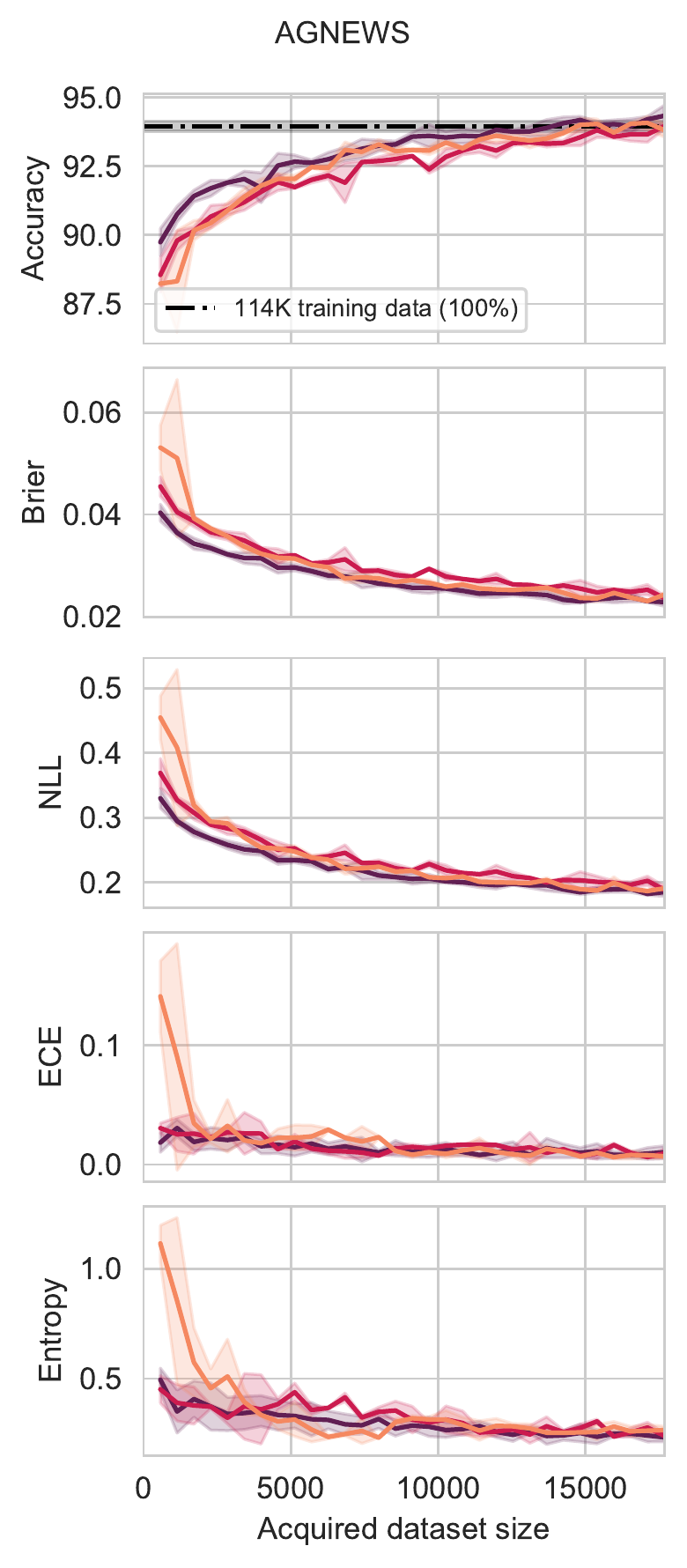}
    \end{subfigure}%
    \\
    \begin{subfigure}{0.9\columnwidth}
        \centering
        \includegraphics[width=0.45\textwidth]{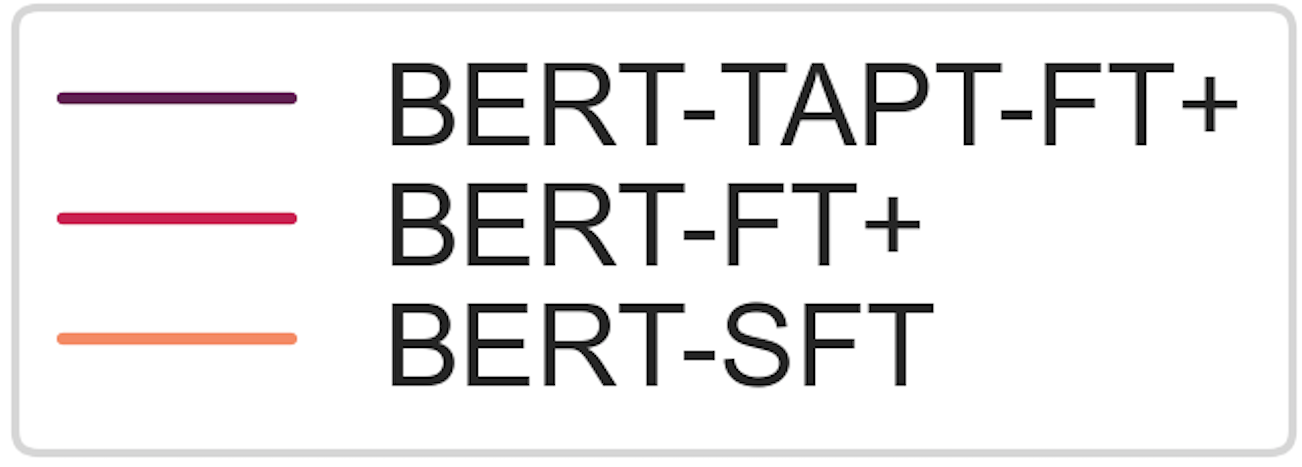}
    \end{subfigure}
    \caption{Ablation study for \tapt{} and \ft{}.}
    \label{fig:early20_vs_baseline}
\end{figure}

Figure~\ref{fig:early20_vs_baseline} shows the results for the smallest and largest datasets, \trec{} and \ag{} respectively.
For \trec{}, 
training \bert{} with our fine-tuning approach \ft{} provides large gains both in accuracy and uncertainty calibration, showing the importance of fine-tuning the LM for a larger number of epochs in low resource settings.
For the larger dataset, \ag{}, we see that \bert{} with \sft{} performs equally to \ft{} which is the ideal scenario. We see that our fine-tuning approach does not deteriorate the performance of \bert{} given the large increase in warmup steps, showing that our simple strategy provides robust results in both high and low resource settings.
After demonstrating that \ft{} yields better results than \sft{}, we next compare \bert-\tapt{}-\ft{} against \bert{}-\ft{}. We observe that in both datasets \bert-\tapt{} outperforms \bert{}, with this being particularly evident in the early iterations.
This confirms our hypothesis that by implicitly using the entire pool of unlabeled data for extra pretraining (\tapt{}), we boost the performance of the AL model using less data.

\section{Conclusion}
We have presented a simple yet effective training scheme for AL with pretrained LMs that accounts for 
varying data availability and instability of fine-tuning.
Specifically, we propose to first continue pretraining the LM
with the available unlabeled data to \textit{adapt} it to
the task-specific domain. This way, we leverage not only the available labeled data at each AL iteration, but the entire unlabeled pool.
We further propose a method to \textit{fine-tune} the model during AL iterations so that training is robust in both low and high resource data settings.

Our experiments show that our approach yields substantially better results than standard fine-tuning in five standard NLP datasets.
Furthermore, we find that \textit{the training strategy can be more important than the acquisition strategy}. In other words, a poor training strategy can be a crucial impediment to the effectiveness of a good acquisition function, and thus limit its effectiveness (even over random sampling). Hence, our work highlights how critical it is to properly adapt a pretrained LM to the low data resource AL setting.

As state-of-the-art models in NLP advance rapidly, in the future we would be interested in exploring the use of larger LMs, such as \textsc{Gpt-3}~\cite{NEURIPS2020_1457c0d6} and \textsc{Flan}~\cite{wei2022finetuned}. These models have achieved impressive performance in very low data resource settings (e.g. zero-shot and few-shot), so we would imagine they would be good candidates for the challenging setting of active learning.

\section*{Acknowledgments}
We would like to thank Giorgos Vernikos, our colleagues at the Sheffield NLP group for feedback on an earlier version of this paper, and all the anonymous reviewers for their constructive comments. KM and NA are supported by Amazon through the Alexa Fellowship scheme.

\bibliography{custom}
\bibliographystyle{acl_natbib}

\clearpage 
\appendix

\section{Appendix: Experimental Setup}\label{sec:appendix_exp_setup}

\subsection{Datasets}\label{sec:datasets}
We experiment with five diverse 
natural language understanding tasks
including binary and multi-class labels and varying dataset sizes (Table~\ref{table:datasets}). 
The first task is question classification using the six-class version of the small \textsc{trec-6} dataset of open-domain, fact-based questions divided into broad semantic categories~\cite{article-trec}.
We also evaluate our approach on sentiment analysis using the binary movie review \textsc{imdb} dataset~\cite{maas-etal-2011-learning} and the binary version of the \textsc{sst-2} dataset~\cite{socher-etal-2013-recursive}.
We finally use the large-scale \textsc{agnews} and \textsc{dbpedia} datasets from ~\citet{NIPS2015_250cf8b5} for topic classification. We undersample the latter and form a $\Dpool$ of $20$K examples and $\Dval$ $2$K as in ~\citet{margatina-etal-2021-active}.
For \trec{}, \imdb{} and \sst{} we randomly sample $10\%$ from the training set to serve as the validation set, while for \ag{} we sample $5$\%. For the \dbpedia{} dataset we undersample both training and validation datasets (from the standard splits) to facilitate our AL simulation (i.e. the original dataset consists of $560$K training and $28$K validation data examples). For all datasets we use the standard test set, apart from the \sst{} dataset that is taken from the \textsc{glue} benchmark~\cite{wang2018glue} we use the development set as the held-out test set (and subsample a development set from the original training set).


\subsection{Training \& AL Details} \label{sec:training_details}
We use \textsc{BERT-base}~\cite{Devlin2019-ou} and fine-tune it 
(\tapt{}~\S\ref{sec:balm_tapt}) for $100$K steps, with learning rate $2e-05$ and the rest of hyperparameters as in \citet{gururangan-etal-2020-dont} using the
HuggingFace library~\cite{wolf-etal-2020-transformers}.
We evaluate the model $5$ times per epoch on $\mathcal{D}_{val}$ and keep the one with the lowest validation loss as in~\citet{Dodge2020FineTuningPL}.
We use the code provided by \citet{Kirsch2019-lk} for the uncertainty-based acquisition functions and \citet{yuan2020coldstart} for \alps{}, \badge{} and \bkm{}.
We use the standard splits provided for all datasets, if available, otherwise we randomly sample a validation set. We test all models on a held-out test set.
We repeat all experiments with five different random seeds resulting into different initializations of $\Dlab$ and the weights of the extra task-specific output feedforward layer.
For all datasets we use as budget the $15\%$ of $\Dpool$.
Each experiment is run on a single Nvidia Tesla V100 GPU.


\subsection{Hyperparameters}\label{sec:hyperparameters} 
For all datasets we train \textsc{BERT-base}~\cite{Devlin2019-ou} from the HuggingFace library~\cite{wolf-etal-2020-transformers} in Pytorch~\cite{NEURIPS2019_9015}. We train all models with batch size $16$, learning rate $2e-5$, no weight decay, AdamW optimizer with epsilon $1e-8$. For all datasets we use maximum sequence length of $128$, except for \imdb{} and \ag{} that contain longer input texts, where we use $256$. To ensure reproducibility and fair comparison between the various methods under evaluation, we run all experiments with the same five seeds that we randomly selected from the range $[1,9999]$. 

\subsection{Baselines}\label{sec:baselines}

\paragraph{Acquisition functions} We compare \entropy with four baseline acquisition functions. 
The first is the standard AL baseline, \textbf{\textsc{Random}}, 
which applies uniform sampling and selects $k$ data points from $\Dpool$ at each iteration. 
The second is 
\textbf{\badge{}}~\cite{Ash2020Deep}, an acquisition function that aims to combine diversity and
uncertainty sampling. 
The algorithm computes \textit{gradient embeddings} $g_x$ for every candidate data point $x$ in $\Dpool$ and then uses clustering to select a batch. Each $g_x$ is computed as the gradient of the cross-entropy loss with respect to the parameters of the model’s last layer.
We also compare against a recently introduced cold-start acquisition function called \textbf{\textsc{Alps}}~\cite{yuan2020coldstart}. 
\textsc{Alps} acquisition uses the masked language model (MLM) loss of \bert{} 
as a proxy for model uncertainty in the downstream classification task. Specifically, aiming to leverage both uncertainty and diversity, \alps{}
forms a \textit{surprisal embedding} $s_x$ for each $x$, by passing the unmasked input $x$ through the \bert{} MLM head
to compute the cross-entropy loss for a random 15\% subsample of tokens against the target labels. 
\alps{} clusters these embeddings to sample $k$ sentences for each AL iteration. Last, following~\citet{yuan2020coldstart}, we use \textbf{\bkm{}} as a diversity baseline, where the $l_2$ normalized \bert{} output embeddings are used for clustering.

\paragraph{Models \& Fine-tuning Methods} We evaluate two variants of the pretrained language model; the original \textbf{\textsc{Bert}} model, used in \citet{yuan2020coldstart} and \citet{Ein-Dor2020-mm}\footnote{\citet{Ein-Dor2020-mm} evaluate various acquisition functions, including entropy with MC dropout, and use \bert{} with the standard fine-tuning approach (\sft{}).},
and our adapted model \textbf{\textsc{Bert}-\textsc{tapt}} (\S\ref{sec:balm_tapt}),
%
%
and two fine-tuning methods;
our proposed fine-tuning approach \textbf{\ft{}} (\S\ref{sec:balm_ft}) and standard \bert{} fine-tuning \textbf{\sft{}}.

\setlength{\tabcolsep}{6pt} 
\renewcommand{\arraystretch}{1.1} 

\begin{table}[h]
\resizebox{\columnwidth}{!}{%
\centering

\begin{tabular}{lcccccc}
\Xhline{2\arrayrulewidth}

\textsc{Model} & \textsc{trec-6} & \textsc{dbpedia}  & \textsc{imdb}& \textsc{sst-2}& \textsc{agnews}\\\hline 
\multicolumn{6}{c}{\textsc{Validation Set}}\\\hline
\textsc{bert} &94.4 & 99.1& 90.7& 93.7& 94.4\\
\textsc{bert-tapt} & 95.2 & 99.2 &91.9& 94.3 & 94.5  \\ \hline
\multicolumn{6}{c}{\textsc{Test Set}}\\\hline
\textsc{bert} & 80.6& 99.2& 91.0& 90.6& 94.0\\
\textsc{bert-tapt} & 77.2 & 99.2 & 91.9& 90.8& 94.2\\ 
\Xhline{2\arrayrulewidth}
\end{tabular}
}

\caption{Accuracy with $100\%$ of data over five runs (different random seeds).} 

\label{table:tapt_100}
\end{table}

\section{Appendix: Analysis}
\subsection{Task-Adaptive Pretraining (\tapt{}) \& Full-Dataset Performance}\label{sec:appendix_tapt}
\begin{figure}[!t]
    \resizebox{\columnwidth}{!}{%
    \begin{subfigure}{\textwidth}
        \centering
        \includegraphics[width=\textwidth]{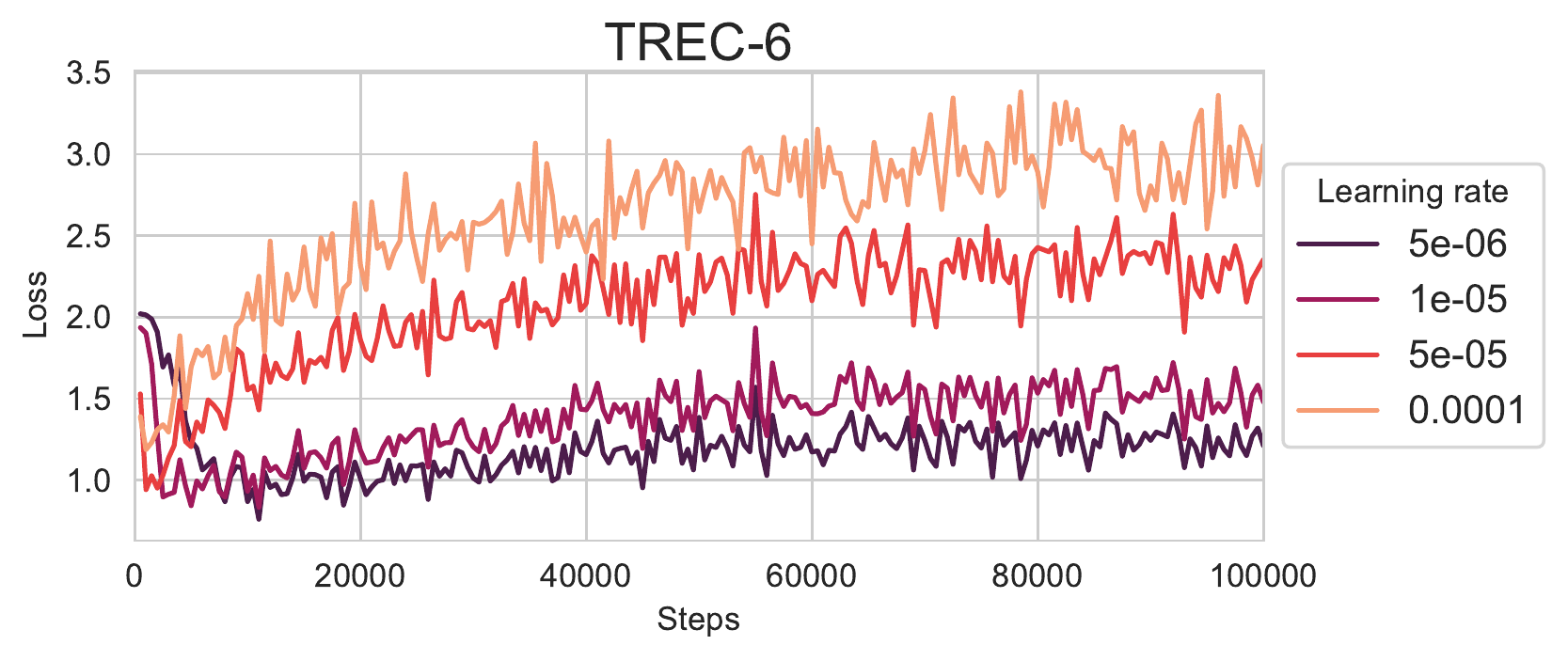}

        \centering
        \includegraphics[width=\textwidth]{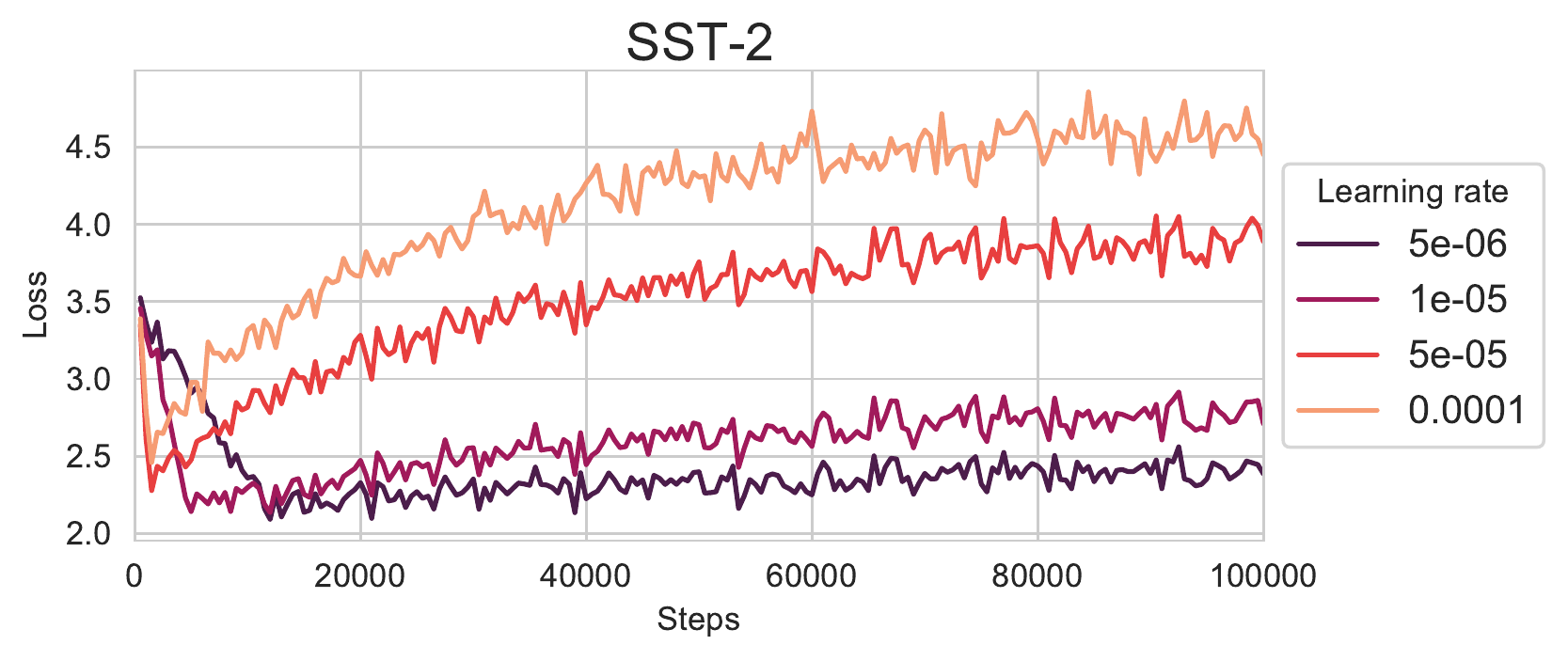}
        
        \includegraphics[width=\textwidth]{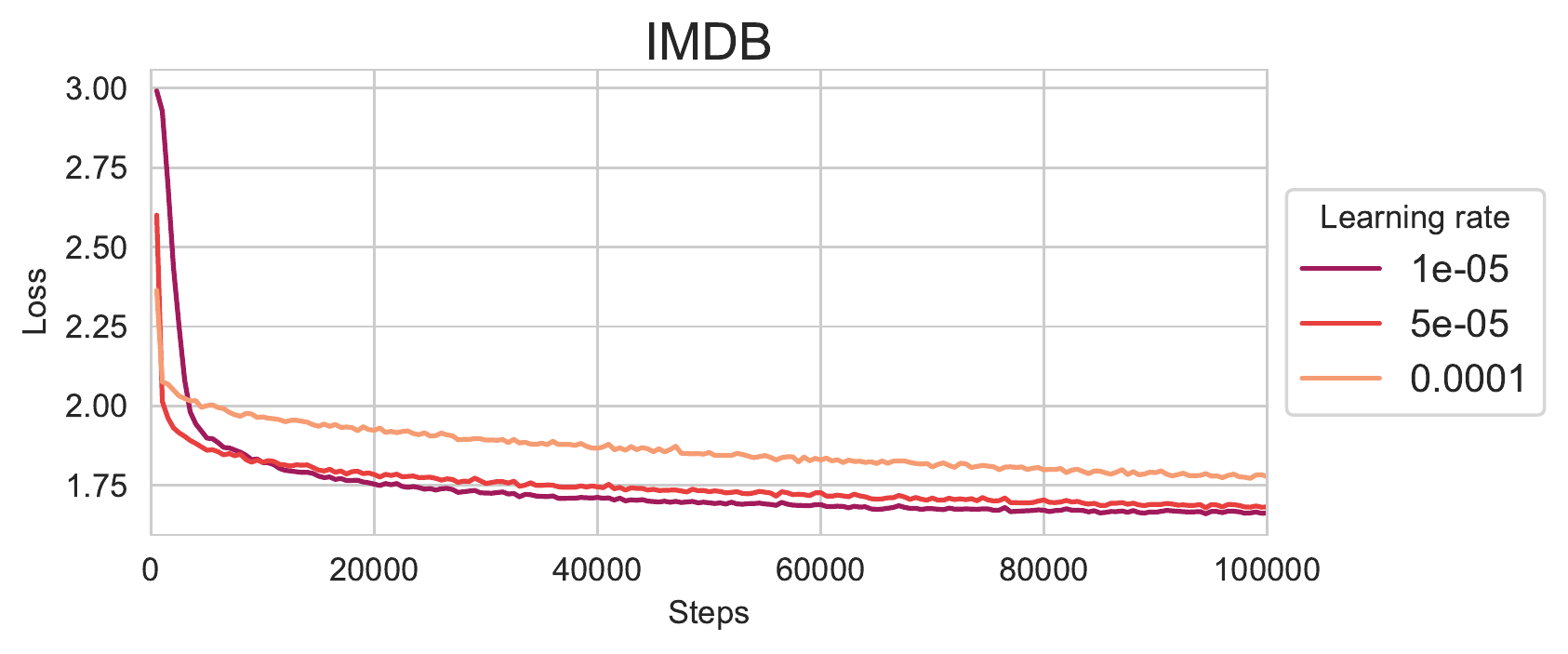}

    \end{subfigure}%
    }
    \caption{Learning curves of \tapt{} for various learning rates.}
    \label{fig:tapt_full}
\end{figure}

As discussed in \S\ref{sec:balm_tapt} and \S\ref{sec:analysis}, we continue training the \textsc{BERT-base}~\cite{Devlin2019-ou} pretrained masked language model using the available data $\Dpool$. We explored various learning rates between $1e-4$ and $1e-5$ and found the latter to produce the lowest validation loss. We trained each model (one for each dataset) for up to $100$K optimization steps, we evaluated on $\Dval$ every $500$ steps and saved the checkpoint with the lowest validation loss. We used the resulting model in our (\bert{}-\tapt{}) experiments.
We plot the learning curves of masked language modeling task (\tapt{}) for three datasets and all considered learning rates in Figure \ref{fig:tapt_full}. We notice that a smaller learning rate facilitates the training of the MLM.

In Table~\ref{table:tapt_100} we provide the validation and test accuracy of \bert{} and \bert-\tapt{} for all datasets. We present the mean across runs with three random seeds. For fine-tuning the models, we used the proposed approach \ft{} (\S\ref{sec:balm_ft}).

\setlength{\tabcolsep}{6pt} 
\renewcommand{\arraystretch}{1.1} 

\begin{table*}[!t]
\centering
\begin{tabular}{lccccc}
\Xhline{2\arrayrulewidth}

\textsc{} & \trec{} & \sst{} & \imdb{} & \dbpedia{} & \ag{}\\\hline 

\rand{}  & 0/0 & 0/0 & 0/0 &  0/0 & 0/0 \\ 
\alps{}  & 0/57 & 0/478 & 0/206 &  0/134 & 0/634\\
\badge{}  & 0/63 & 0/23110 & 0/1059& 0/192& -\\
\bkm{}  & 0/47 & 0/2297 & 0/324 &  0/137 & 0/3651\\
\entropy{} & 81/0 & 989/0 & 557/0 & 264/0 &2911/0 \\
\lc{} & 69/0 & 865/0 & 522/0 & 256/0 & 2607/0 \\
\bald{} & 69/0 & 797/0 &524/0 &256/0 & 2589/0 \\
\bb{} & 69/21 & 841/1141 & 450/104 & 256/482 & 2844/5611 \\\hline

\Xhline{2\arrayrulewidth}
\end{tabular}

\caption{Runtimes (in seconds) for all datasets. In each cell of the table we present a tuple $i / s$ where $i$ is the \textit{inference time} and $s$ the \textit{selection time}. \textit{Inference time} is the time for the model to perform a forward pass for all the unlabeled data in $\Dpool$ and \textit{selection time} is the time that each acquisition function requires to rank all candidate data points and select $k$ for annotation (for a single iteration). Since we cannot report the runtimes for every model in the AL pipeline (at each iteration the size of $\Dpool$ changes), we provide the median.}

\label{table:runtimes}
\end{table*}

\begin{figure}[!h]
    \begin{subfigure}{0.45\columnwidth}
        \centering
        \includegraphics[width=\textwidth]{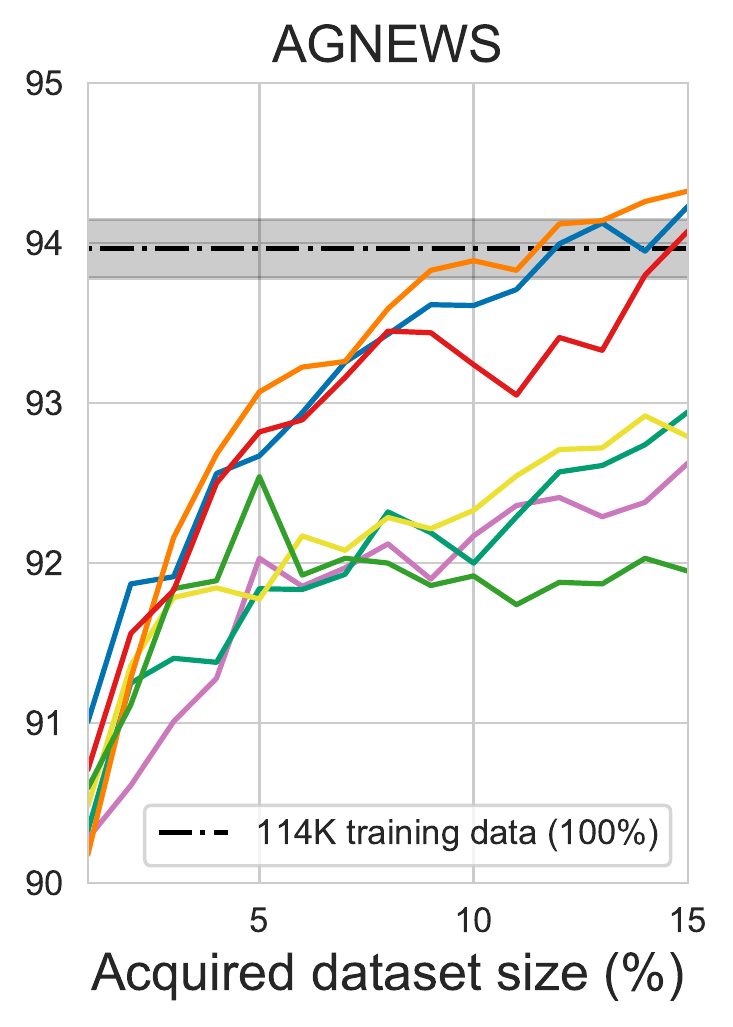}
    \end{subfigure}%
    \hfill
    \begin{subfigure}{0.45\columnwidth}
        \centering
        \includegraphics[width=\textwidth]{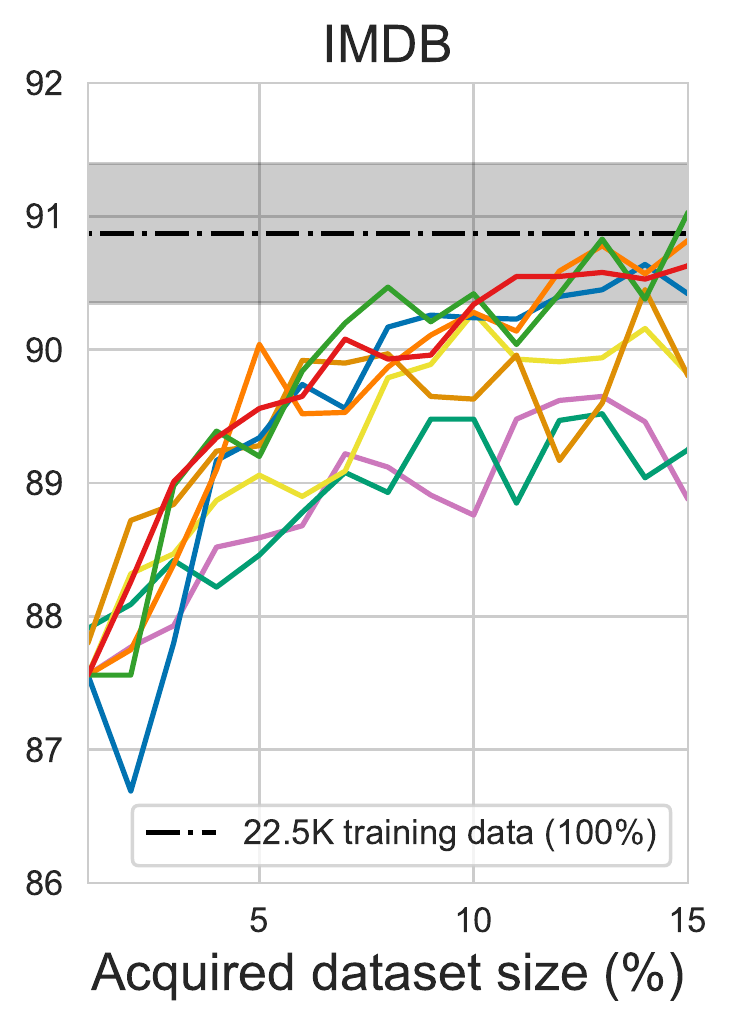}
    \end{subfigure}%
    \\
    \begin{subfigure}{0.9\columnwidth}
        \centering
        \includegraphics[width=0.45\textwidth]{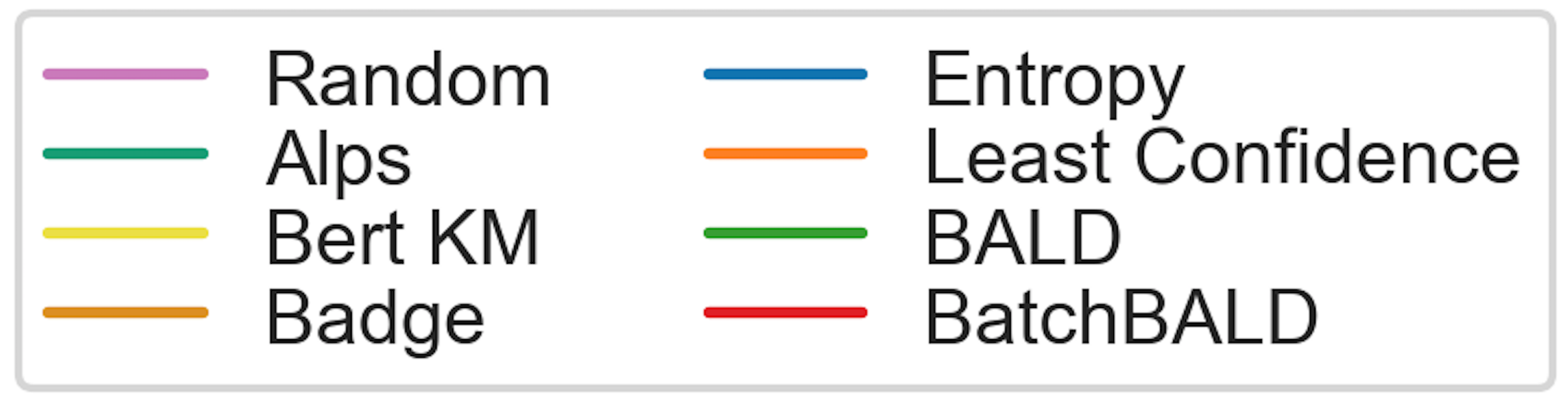}
    \end{subfigure}
    \caption{Comparison of acquisition functions using \tapt{} and \ft{} in training \bert{}.}
    \label{fig:all_afs}
\end{figure}
\subsection{Performance of Acquisition Functions}\label{sec:perf_afs}
In our \bert{}-\tapt-\ft{} experiments so far, we showed results with \entropy{}. We have also experimented with various uncertainty-based acquisition functions. Specifically, 
four uncertainty-based 
acquisition functions are 
used
in our work:
\lc{}, \entropy{}, \bald{} and \bb{}.
\lc{}~\cite{Lewis:1994:SAT:188490.188495} sorts $\mathcal{D}_{pool}$ by the probability of \textit{not} predicting the most confident class, in descending order,
\entropy{}~\cite{Shannon1948} selects samples that maximize the predictive entropy,
and
\bald{}~\cite{Houlsby2011-qz}, short for Bayesian Active Learning by Disagreement,  chooses data points that maximize the mutual information between predictions and model's posterior probabilities.
\bb{}~\cite{Kirsch2019-lk} is a recently introduced extension of BALD that \textit{jointly} scores points by estimating the mutual information between multiple data points and the model parameters. This iterative algorithm aims to find \textit{batches} of informative data points, in contrast to \bald{} that chooses points that are informative individually.
Note that \lc{}, \entropy{} and \bald{} have been used in AL for NLP by \citet{Siddhant2018-lg}. To the best of our knowledge, \bb{} is evaluated for the first time in the NLP domain.

Instead of using the output softmax probabilities for each class, we use a probabilistic formulation of deep neural networks in order to acquire better calibrated scores.
Monte Carlo (MC) dropout~\cite{Gal2016-lf} is a simple yet effective method for performing approximate variational inference, based on dropout~\cite{Srivastava2014-bs}.
\citet{Gal2016-lf} prove that by simply performing \textit{dropout during the forward pass in making predictions}, the output is equivalent to the prediction when the parameters are sampled from a variational distribution of the true posterior. Therefore, dropout during inference results into obtaining predictions from different parts of the network.
Our \bert-based $\mathcal{M}_i$ model uses dropout layers during training for regularization.
We apply MC dropout by simply activating
them during test time and we perform multiple stochastic forward passes. 
Formally, we do $N$ passes of every $x \in \Dpool$ through $\mathcal{M}_i(x;W_i)$ to acquire $N$ different output probability distributions for each 
$x$.
MC dropout for AL has been previously used in the literature~\cite{Gal2017-gh, Shen2017-km, Siddhant2018-lg, Lowell2019-mf, Ein-Dor2020-mm,shelmanov-etal-2021-active}.

Our findings show that all functions provide similar performance, except for \bald{} that slightly underperforms.
This makes our approach agnostic to the selected uncertainty-based
acquisition method.
We also evaluate our proposed methods with our baseline acquisition functions, i.e. \rand{}, \alps{}, \bkm{} and \badge{}, since our training strategy is orthogonal to the acquisition strategy. We compare all acquisition functions with \bert-\tapt-\ft{} for \ag{} and \imdb{} in Figure~\ref{fig:all_afs}. We observe that in general uncertainty-based acquisition performs better compared to diversity, while all acquisition strategies have benefited from our training strategy (\tapt{} and \ft{}).

\subsection{Efficiency of Acquisition Functions}\label{sec:appendix_acquisition}
In this section we discuss the efficiency of the eight acquisition functions considered in this work; \rand{}, \alps{}, \badge{}, \bkm{}, \entropy{}, \lc{}, \bald{} and \bb{}.

In Table~\ref{table:runtimes} we provide the runtimes for all acquisition functions and datasets. 
Each AL experiments consists of multiple iterations and (therefore multiple models), each with a different training dataset $\Dlab$ and pool of unlabeled data $\Dpool$. 
In order to evaluate how computationally heavy is each method, we provide the \textit{median} of all the models in one AL experiment. We calculate the runtime of two types of functionalities. The first is the \textit{inference time} and stands for the forward pass of each $x \in \Dpool$ to acquire confidence scores for uncertainty sampling. \rand{}, \alps{}, \badge{} and \bkm{} do not require this step so it is only applied of uncertainty-based acquisition where acquiring uncertainty estimates with MC dropout is needed. The second functionality is \textit{selection time} and measures how much time each acquisition function requires to rank and select the $k$ data points from $\Dpool$ to be labeled in the next step of the AL pipeline. \rand{}, \entropy{}, \lc{} and \bald{} perform simple equations to rank the data points and therefore so do not require selection time. On the other hand, \alps{}, \badge{}, \bkm{} and \bb{} perform iterative algorithms that increase selection time. From all acquisition functions \alps{} and \bkm{} are faster because they do not require the inference step of all the unlabeled data to the model. \entropy{}, \lc{} and \bald{} require the same time for selecting data, which is equivalent for the time needed to perform one forward pass of the entire $\Dpool$. Finally \badge{} and \bb{} are the most computationally heavy approaches, since both algorithms require multiple computations for the \textit{selection time}. \rand{} has a total runtime of zero seconds, as expected.

\end{document}